\pdfoutput=1
\documentclass[11pt,table]{article}
\usepackage[]{acl}
\usepackage{times}
\usepackage{latexsym}
\usepackage[T1]{fontenc}
\usepackage[utf8]{inputenc}
\usepackage{microtype}
\usepackage{inconsolata}
\usepackage{tabularx}
\usepackage{graphicx}
\usepackage{amsmath,amsfonts,amssymb}
\usepackage{booktabs}
\usepackage{scalefnt}
\usepackage{balance}
\usepackage{xcolor}
\usepackage{paralist}
\usepackage{multirow}
\renewcommand{\paragraph}[1]{\noindent\textbf{#1}}
\title{``You are an expert annotator'':\\ Automatic Best--Worst-Scaling Annotations for Emotion Intensity Modeling}

\author{Christopher Bagdon$^{1,3,4}$, Prathamesh Karmalker$^{2}$,\\\bfseries Harsha Gurulingappa$^{2}$, \and Roman Klinger$^{4}$ \\
  $^{1}$Merck Data \& AI Organization, Merck Group, Darmstadt, Germany\\
  $^{2}$Merck Data \& AI Organization, Merck IT Centre, Merck Group, Bengaluru, India\\
  $^{3}$Institut f\"ur Maschinelle Sprachverarbeitung, University of Stuttgart, Germany\\
  $^{4}$Fundamentals of Natural Language Processing, University of Bamberg, Germany\\
    \texttt{\{christopher.bagdon,roman.klinger\}@uni-bamberg.de}\\
  \texttt{\{prathamesh.karmalkar,harsha.gurulingappa\}@merckgroup.com}
  }

\begin{document}
\maketitle
\begin{abstract}
  Labeling corpora constitutes a bottleneck to create models for new
  tasks or domains. Large language models
  mitigate the issue with automatic corpus labeling methods,
  particularly for categorical annotations. Some NLP tasks such as
  emotion intensity prediction, however, require text regression, but
  there is no work on automating annotations for continuous label
  assignments. Regression is considered more challenging
  than classification: The fact that humans perform worse when tasked
  to choose values from a rating scale lead to comparative annotation
  methods, including best--worst scaling. This raises the question if
  large language model-based annotation methods show similar patterns,
  namely that they perform worse on rating scale annotation tasks than
  on comparative annotation tasks. To study this, we automate emotion
  intensity predictions and compare direct rating scale predictions,
  pairwise comparisons and best--worst scaling. We find that the
  latter shows the highest reliability. A transformer regressor
  fine-tuned on these data performs nearly on par with a model trained
  on the original manual annotations.
\end{abstract}

\section{Introduction}
Labeling data with trained experts or via crowdsourcing is a
resource-intensive and time-consuming process
\citep{zaidan2012crowdsourcing,GPT3-labeling,Bunte2021WhyII}.  This
motivates automated annotation methods, including weak supervison
\citep{Snorkel}, zero-shot predictions \citep{GPT3}, or, more
recently, generative models \cite{GPT3-labeling, Cost-Eff-Anno,GPT3}.
Depending on the downstream task at hand, the labels to be assigned to
a textual instance are categorical (e.g., in text classification),
structured (for instance in parsing or named entity recognition), or
continuous (e.g., emotion intensity, sentiment strength, or
personality profiling predictions).

Annotating for continuous value labels comes with its own set of
challenges. It can be difficult to obtain consistent labels from
humans by asking them to assign a value from a rating scale
\cite{schuman-likert}. Not only is it difficult for the annotator to 
rate texts consistently, but it is also difficult for researchers to 
design rating scales, as there are many design decisions which can 
bias the annotator, such as scale point descriptions and scale 
granularity. This lead to comparative annotation setups, in
which annotators are tasked to compare multiple instances for the same
task, which is easier to accomplish and has fewer design decisions to make.
Consider the two example sentences for sentiment strength:
\begin{quote}
  (1) She's quite happy.\\
  (2) He is extremely delighted.
\end{quote}
It is difficult to assign a value $v(s_i)\in[-1;1]$ to these
sentences in isolation, or even in context, but it is straight-forward
to decide that $v(s_2)>v(s_1)$.

Best--worst scaling \cite[BWS,][]{Louv-Food-BWS,Louv-BWS} is
such a comparison-based annotation method that has proven to be more
reliable than assigning values on rating scales, when annotating
continuous values. The idea is to task an annotator to decide which
instance is the one with the highest and the lowest value. The tuple
size can be varied but \newcite{BWS-reliable}
observe that quadruples provide a good trade-off between context and
numbers of comparative judgements.

In this paper, we question if BWS is also an appropriate approach for
large language model-based annotations. On the one hand, one might
argue that comparative tasks are also more reliably conducted with
language models. On the other hand, one might argue that a large
language model (LLM) has more access to other text instances
implicitly from training data which it can compare a text to, than a
human. That would be an argument that when using a LLM for annotations, BWS may not be
necessary.

We therefore set up prompts for continuous value assignments for two
direct and two comparison-based annotation approaches. We use rating
scales \citep[RS,][]{likert1932technique}, in which the model directly
labels texts with a numerical value, in two variants: annotating
single texts and tuples of four texts per prompt. Our comparison-based
approaches are paired comparisons \citep[PC,][]{PairCompare} in which
every text is compared to every other text, and best--worst scaling
\citep[BWS,][]{Louv-BWS}, where ``best'' and ``worst'' instances are
picked from a tuple.

In our evaluation, we focus on comparisons against human annotations
for the emotion intensity prediction task \cite{AIT-Main}. We compare
the LLM-based annotations directly against human annotations and
further train a transformer-based regressor both on the human-labeled
data and the LLM-labeled data. The motivation for this regressor is to
avoid the requirement to put together instances in tuples and query a
potentially expensive API at inference time.
We answer the following research questions:

\begin{compactenum}
\item Does the best--worst scaling annotation method perform better
  than rating scales or paired comparisons when using generative
  models for labeling text with continuous values? \textit{(Yes, it
    does.)}
\item How does the performance of a transformer-based regressor
  compare when trained with automated annotations vs.\ human
  annotations? \textit{(The models perform on par.)}
\end{compactenum}

\section{Related Work}
\subsection{Automated Annotation}
Annotating texts can be an arduous and costly task
\citep{zaidan2012crowdsourcing}. Crowd-sourced annotation is sometimes
cheaper than following a more traditional approach to hire few expert
annotators \citep{snow-etal-2008-cheap}, but generally, the costs increase
with the difficulty of the task, either because more careful training
is needed or more annotators need to be involved to obtain reliable
aggregated scores.

This situation lead to the development of automatic annotation
methods. \textit{Weak supervision} uses noisy automated annotations to
train models. The expectation is that they might be less accurate than
supervised models, but still better than unsupervised learning
\citep{Snorkel}. Examples include the use of heuristics, keyword
searches, or distant supervision from databases \citep{Snorkel,
  distant_super, keyword_weak}.  A more recent approach to automatic
data labeling is zero-shot classification \citep{GPT3}. It relies on
the information present in a pretrained language model to solve the
task, either via textual entailment \citep{yin-etal-2019-benchmarking,
  zero-relation-entail, plaza-del-arco-etal-2022-natural}, by mapping
verbalizations of classes to outputs of autoregressive models
\citep[i.a.]{shin-etal-2020-autoprompt}, or with instruction-tuned
models
\citep[i.a.]{zhang-etal-2023-aligning,gupta-etal-2022-instructdial,ghosh-etal-2023-lasque}.

Recently, \cite{wadhwa2023using} successfully used rating scales with
LLM's to improve crowdsourced annotations. Their goal was, however,
not automatic annotations but improving existing annotations.

\subsection{Annotating Continuous Values}
Some NLP tasks require the prediction of continuous values, for
instance rating emotion intensity \citep{EmoInt17}. A typical
operationalization is to ask annotators to choose a position on a
rating scale, such as Likert scales \citep{likert1932technique}. The
exact position that humans chose does, however, depend on various
subjective aspects, including preferences for particular intervals
\citep{BWS-reliable}. This leads to inconsistencies between
annotators. Annotations can also be inconsistent from a single
annotator -- after seeing more examples they might adjust their
interpretation of a specific value range.

An alternative is to rate data-points via comparison such
as Paired Comparisons \citep[PC,][]{PairCompare}. PC ranks
texts by comparing every text to every other text. This approach comes
with the major drawback of a quadratic number of required annotator's
decisions in the number of instances.

\begin{figure*}
    \centering
    \includegraphics[width=0.85\linewidth]{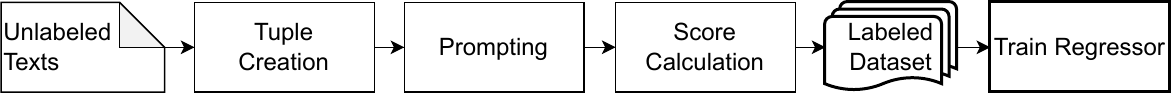}
    \caption{Workflow diagram of our method to use automated annotations for training a regression model.}
    \label{fig:Flow}
\end{figure*}

\subsection{Best--Worst Scaling}
\label{sec:BWS}
Best--worst Scaling (BWS) addresses the issues of rating scales and
paired comparisons \citep{Louv-BWS, Louv-Food-BWS}. Annotators are
provided an $n$-tuple (typically $n=4$) of data-points and asked to
choose the \textit{best} and \textit{worst} of the tuple, given the
scale they are annotating. 4-tuples are efficient because by giving
the \textit{best} and \textit{worst} ratings the annotator has
effectively done five out of six possible pair-wise comparisons.
\newcite{BWS-reliable} showed that $1.5N$ to $2N$ annotations, where
$N$ is the number of instances, lead to reliable results. The final
score $s(i)$ of an instance $i$ is computed by
$s(i)=\frac{\#\mathrm{best}(i)-\#\mathrm{worst}(i)}{\#\mathrm{overall}(i)}$.

\citet{BWS-reliable} found BWS produces more consistent annotations
than RS, while also being efficient in the number of annotations
required. They compared annotations from RS and BWS using split-half
reliability (SHR). As qualitative values cannot be precisely known,
one measure of their accuracy is reproducibility across multiple
annotators. SHR evaluates reproducibility by splitting all annotations
for each data-point randomly into two bins. Each bin is used to
calculate the target label separately, and the two sets of scores are
compared. This is repeated for multiple iterations and the
correlation scores are averaged.

\subsection{Emotion Intensity}
Emotion analysis consists of various subtasks, including emotion
categorization \citep[i.a.]{Calvo2013,AIT18}, in
which emotion labels from a predefined set are assigned. The labels
commonly stem from psychological models of basic emotions
\citep{Plutchik2001}. Another popular task in emotion analysis is
affect prediction, where continuous values of
valency and arousal are assigned \citep{Posner2005}.

A combination of these approaches in which intensities for a given
emotion are to be predicted was first proposed with a shared task
in 2017 \citep{EmoInt17}. The Affect-in-Tweets Dataset is an extension
of the original data \cite[AIT,][]{AIT-Main}.\footnote{More
  information, including terms and conditions, can be found at
  \url{https://competitions.codalab.org/competitions/17751}.} They
have been created via best--worst scaling with a setup in which
annotators have been asked to select the most intense and least
intense instance from a quadruple of tweets for a given emotion. The
data set is partitioned into the four emotions joy, fear, anger, and
sadness. An individual instance can appear in multiple of the data
sets.

\section{Methods}
We now provide an overview of our automatic annotation method
(\S\ref{sec:overview}), describe our prompting strategies
(\S\ref{sec:annotation}), and finally explain how the automatically
created corpora are used to estimate a transformer-based regressor
(\S\ref{sec:regressor}).

\subsection{Overview}
\label{sec:overview}
Our method automates the annotations of a training set which we use to
estimate a regressor. Figure \ref{fig:Flow} shows the workflow,
discussed in the following.

\paragraph{Tuple Creation.}
In this step of the pipeline, we create the instances for
annotation. Rating scales (RS) require single instances. For rating
scales tuples (RS-T) we use 4-tuples to give the model the same
context per prompt as BWS, where we also use 4-tuples. Paired
comparisons (PC) require each text to be paired with every other text
once. In BWS, nearly no pair appears in more than one tuple and all
texts appear in nearly the same number of tuples. We follow the
recommendation by \citet{BWS-reliable} to use $1.5N$ to $2N$ tuples.

\paragraph{Prompting.}
The prompt tasks the LLM to output annotations for the given text
tuple. Independent of the concrete prompting approach (RS, RS-T, PC, BWS),
each prompt contains a role description, a task description, the
texts, and formatting instructions. As the back-end, we mainly use
GPT-3.5-turbo\footnote{https://platform.openai.com/docs/models/gpt-3-5} and
compare the best performing setup to GPT-3 and Llama2
\citep{GPT3,touvron2023llama}. We discuss the prompts in
Section~\ref{sec:annotation}.

\paragraph{Score Calculation.}
For RS and RS-T, we directly use the output value as the score. For PC
and BWS, we use the counting method (\S\ref{sec:BWS}). All results are
linearly normalized to $[0;1]$.

\begin{table*}
  \centering\sffamily\small\scalefont{0.85}
  \begin{tabularx}{\linewidth}{p{9mm} p{19mm} | p{19mm} XX}
    \toprule
    Section & Rating Scales & Rating Scales\par Tuples & Paired Comparison & Best--Worst Scaling \\
    \cmidrule(r){1-1}\cmidrule(rl){2-3}\cmidrule(rl){4-4}\cmidrule(rl){5-5}
    Role
    &\multicolumn{2}{p{42.5mm}}{\cellcolor{yellow!33!red!33!blue!20}You are an expert annotator special\-izing in emotion recognition.}
    &\cellcolor{yellow!33!red!33!blue!20}You are an expert annotator specializing in emotion recognition.
    &\cellcolor{yellow!33!red!33!blue!20}You are an expert annotator specializing in emotion recognition.
    \\    \cmidrule(r){1-1}\cmidrule(rl){2-3}\cmidrule(rl){5-5}\cmidrule(rl){4-4}
    Task Descr.
    &\multicolumn{2}{p{42.5mm}}{\cellcolor{yellow!10}Please rate the following text from\par social media for how intense the\par authors feels \{emo\}.}
    &\cellcolor{red!10}Which of the two speakers is likely to be the MOST \{emo\} and which of the two speakers is likely to be the LEAST \{emo\}?
    &\cellcolor{blue!10}Which of the four speakers is likely to be the MOST \{emo\} and which of the four speakers is likely to be the LEAST \{emo\}?
    \\    \cmidrule(r){1-1}\cmidrule(rl){2-3}\cmidrule(rl){5-5}\cmidrule(rl){4-4}
    Scale
    &\multicolumn{2}{p{42.5mm}}{\cellcolor{yellow!10}Use the following scale\par [Round to the fourth decimal.]:\par
    4: extremely intense \{emo\}\par
    3: very intense \{emo\}\par
    2: moderately intense \{emo\}\par
    1: slightly intense \{emo\}\par
    0: Not \{emo\} at all}
    &
    \\    \cmidrule(r){1-1}\cmidrule(rl){2-3}\cmidrule(rl){5-5}\cmidrule(rl){4-4}
    Format
    &\multicolumn{2}{p{42.5mm}}{\cellcolor{yellow!10}Only reply with the\par numerical rating.}
    &\cellcolor{red!50!blue!20}Only give the Speaker number. Do not repeat the text content.
    &\cellcolor{red!50!blue!20}Only give the Speaker number. Do not repeat the text content.
    \\    \cmidrule(r){1-1}\cmidrule(rl){2-3}\cmidrule(rl){5-5}\cmidrule(rl){4-4}
    Texts
    &\cellcolor{yellow!10}Text: \{text\}
    &\cellcolor{yellow!10}Text 1: \{text1\}\par
    Text 2: \{text2\}\par
    Text 3: \{text3\}\par
    Text 4: \{text4\}
    &\cellcolor{red!10}Speaker 1: \{text1\}\par
    Speaker 2: \{text2\}
    &\cellcolor{blue!10}Speaker 1: \{text1\}\par
    Speaker 2: \{text2\}\par
    Speaker 3: \{text3\}\par
    Speaker 4: \{text4\}
    \\    \cmidrule(r){1-1}\cmidrule(rl){2-3}\cmidrule(rl){5-5}\cmidrule(rl){4-4}
    Format\par Example
    &\multicolumn{2}{p{42.5mm}}{\cellcolor{yellow!10}Format your response as:\par \{emo\} intensity:}
    &\cellcolor{red!10}Format your response as:\par
    Most \{emo\} Speaker:\par
    Least \{emo\} Speaker:
    &\cellcolor{blue!10}Format your response as:\par
    Most \{emo\} Speaker:\par
    Least \{emo\} Speaker:
    \\
    \bottomrule
  \end{tabularx}
  \caption{Prompts for Rating Scales, Rating Scales Tuples, Paired
    Comparisons, and Best--Worst Scaling. Variables are typeset in
    \{curly brackets\}. Unique text blocks are in the same color
    across a row. Rounding is only requested for decimal
    scales. Rating Scales and Rating Scales Tuples are identical
    except for the Texts section.}
  \label{tab:prompts}
\end{table*}

\subsection{Annotation}
\label{sec:annotation}
The prompt differs for each of the four annotation methods. We show
all of them side-by-side in Table~\ref{tab:prompts}. Each prompt
contains up to six parts:
The \textit{role} informs the model how we expect it to behave. For
GPT-3.5, this is applied as a system level prompt separate from the
main prompt. With models which do not utilize a system prompt, it is
the first line of the prompt.
The \textit{task description} is similar to the instructions for humans.
The \textit{scale} is only used for RS and RS-T, and the scale in Table~\ref{tab:prompts} is only an example. It is updated according to
the actual rating scale that is used.
The \textit{format} explains the expected output based on the
\textit{texts} that are labeled so that the model can refer
to them.
The final element is an \textit{example} for the expected output.

\paragraph{Rating Scales.}
For rating scales, which directly annotate individual texts, we do not
need tuples. The scale, which we consider a parameter of this method,
is included as part of the task description. When the scale includes
decimals we instruct the model to round to the fourth digit.

\paragraph{Rating Scales Tuples.}
Comparison-based approaches have the advantage of showing the model more examples
of text from the dataset per prompt. To mitigate this advantage we have the model rate four texts per prompt. Otherwise, this approach is identical to our rating scale approach. 

\paragraph{Paired Comparisons.}
Paired Comparisons compare every text with every other text. We
instruct the model to choose both the speaker with the \textit{most}
emotion and the \textit{least} emotion. We use the terminology
``speaker'' to follow AIT's task description \citep{AIT-Main}. We
accept an output if it includes two distinct predictions.

\paragraph{Best--worst Scaling.}
Our setup of the BWS annotation process follows
\citet{BWS-reliable,AIT-Main}.  We use tuples of four texts, with no
pair of texts appearing in more than one tuple. The tuples are
annotated one at a time, though a single prompt is used to annotate
both \textit{most} and \textit{least}. We accept an output if it
includes two distinct predictions. We request an output for the same
tuple multiple times until we receive an acceptable
answer.\footnote{In our experiments with GPT3.5 (200,000 tuple
  requests), few non-acceptable answers have been returned (388). The
  first repetition typically lead to an acceptable answer.}

\subsection{Regressor}
\label{sec:regressor}
In principle, the output of a LLM can directly be used to label unseen
instances at application time. Nevertheless, we consider it reasonable
to train a regressor on top of the annotations for three
reasons. Firstly, it makes the use of the annotations 
comparable to human annotations which are also the input to a model
training. Secondly, it allows the smaller regressor to run locally and
does not require a potentially expensive API to be called, that might
also change in behaviour. Thirdly, for PC and BWS, annotating in a
zero-shot learning setting at inference time would require combining
the instance of interest in tuples.
We fine-tune roberta-base \citep{liu2019roberta} with a regression
head with default parameters for 5
epochs.\footnote{\url{https://huggingface.co/transformers/v3.0.2/model_doc/auto.html},
  we tested other epoch counts, including early stopping, in
  preliminary experiments but did not find the results to differ
  substantially.}

\section{Experiments}
We perform experiments to compare the annotation methods on the
emotion intensity prediction task, firstly to understand if BWS, PC,
RS, or RS-T perform differently and secondly to understand their performance in
comparison to human annotators. Finally, we perform additional
experiments to understand the role of the tuple count, which can be
flexibly adapted in automated annotations.

\subsection{General Setup}

\paragraph{Dataset.}
We use the Affect-in-Tweets Dataset (AIT). AIT has been manually
annotated for the 2018 shared task \textit{SemEval-2018 Task 1: Affect
  in Tweets} \citep{AIT18}.
It consists of tweets, manually annotated using BWS for emotion
intensity scores for joy, sadness, anger, and fear. AIT is an
extension of the Emotion-Intensity dataset (EmoInt) \citet{EmoInt17};
its training set is composed of the entirety of EmoInt and its
development and test sets are newly added. Table~\ref{tab:datastats}
shows the statistics.

\begin{table}
  \centering\small
  \begin{tabular}{lrrr}
    \toprule
    \multicolumn{1}{c}{Emotion} & Training& Dev& Test\\
    \cmidrule(r){1-1} \cmidrule(lr){2-2} \cmidrule(lr){3-3}\cmidrule(l){4-4}
    Joy&1,616& 290&1,105\\
    Anger& 1,701& 388&1,002\\
    Fear& 2,252& 389&986\\
    Sadness& 1,533& 397&975\\
    \cmidrule(r){1-1}\cmidrule(lr){2-2} \cmidrule(lr){3-3}\cmidrule(l){4-4}
    Total& 7,102& 1,464&4,068\\ \bottomrule
 \end{tabular}
    \caption{Details of the AIT dataset, of which the training set is made up of the EmoInt dataset.}
    \label{tab:datastats}
\end{table}

In the BWS annotation, each tuple consists of four tweets. The target
emotion for choosing the most and least emotion intensity is
predefined. The total number of tuples created per emotion is $2N$
where $N$ is the number of tweets to be annotated.  Each tuple was
annotated by 3--4 independent workers on the crowd-sourcing platform
Crowdflower. The final scores are calculated as described in
Section~\ref{sec:BWS} and linearly scaled to $[0;1]$.

\paragraph{Evaluation.}
AIT uses split half reliability (SHR) to evaluate its annotations for
reproducibility. This approach is problematic for evaluating our
method; we are not taking measurements from multiple people, we are
taking multiple measurements from the same language model. Evaluating
reproducibility is more a test of the model's consistency than it is a
measure of distance from truth. Hence we compare the model's output
to human-annotated scores.

We use a direct and an indirect, downstream evaluation: (1) in the
direct evaluation we compare the AIT's training data gold annotations
to the annotations from the LLM on the training data. This tells us
how well the generative model is able to replicate manual
annotations. (2) The indirect downstream evaluation is based on the
trained regressor model that produces scores in a second
step. Therefore, it is applied to the official test data set. We
consider this evaluation to be more important as it replicates the
actual use-case of such automatic annotations. We use
Pearson's correlation.

\subsection{RQ 1: Does the best--worst scaling annotation method
  perform better than rating scale-based annotations?}
\subsubsection{Experiment Settings Details}

\paragraph{Generative Model.}
We use GPT-3.5-turbo via the AzureOpenAI API. The models are no
different from those made available directly by OpenAI. However,
AzureOpenAI has additional content filters which prevent prompts
containing violence, bigotry, self-harm, or sexual
content.\footnote{https://learn.microsoft.com/en-us/azure/ai-services/openai/concepts/content-filter}
In the case that such filter prevents an output, we rerun the
prompt-tuple in OpenAI's API. Overall, we paid 125.12 Euro for our
experiments.

\paragraph{Best--worst Scaling.}
We use the exact same tuple sets as used in the original manual
annotation of AIT, kindly provided to us by the authors.

\begin{table*}
  \centering\small
  \newcommand{\cc}[1]{\multicolumn{1}{c}{#1}}
  \begin{tabular}{l rrrrrr rrrrrr cc}
    \toprule
    & \multicolumn{6}{c}{Rating Scales} & \multicolumn{6}{c}{Rating Scales Tuples} & BWS & PC \\ 
    \cmidrule(r){2-7}\cmidrule(lr){8-13}\cmidrule(lr){14-14}\cmidrule(lr){15-15}
    \multicolumn{1}{c}{Emo.}
    & \cc{B-1} & \cc{OL-1}   & \cc{B-10}  & \cc{OL-10}  & \cc{D-4}   & \cc{D-10} & 
    \cc{B-1} & \cc{OL-1}   & \cc{B-10}  & \cc{OL-10}  & \cc{D-4}   & \cc{D-10} & 
    \cc{$2N$} & 200P \\
    \cmidrule(r){1-1}\cmidrule(rl){2-7}\cmidrule(rl){8-13}\cmidrule(rl){14-14}\cmidrule(rl){15-15}
    Joy                  & 5.3  & 5.5  & 49.1 & 49.2 & 48.8 & 58.2 & 67.1 & 66.9 & 40.7 & 65.2 & 65.3 & 64.6
 & 81.0 & 81.2 \\
    Ang.               & 15.0 & 16.6 & 39.7 & 42.4 & 52.7 & 57.3 & 
    68.2 & 68.0 & 70.8 & 67.3 & 69.1 & 69.0 &  74.5 & 72.9 \\
    Fear                 & 16.4 & 15.9 & 54.0 & 55.9 & 63.4 & 65.0 & 52.3 & 50.5 & 65.2 & 51.7 & 60.7 & 63.0 & 76.2 & 75.1\\
    Sad.              & 17.9 & 15.6 & 62.6  & 62.1 & 60.1 & 67.0 & 67.8 & 68.4 & 71.4 & 68.4 & 69.5 & 68.6  & 80.3 & 76.8\\
    \cmidrule(rl){2-7}\cmidrule(rl){8-13}\cmidrule(rl){14-14}\cmidrule(rl){15-15}
    Mean                 & 14.9 & 13.7  & 52.0 & 53.1 & 57.3 & 62.4 & 63.6 & 63.1 & 63.7 & 62.8 & 66.0 & 66.3 & 78.1 & 76.8 \\ \bottomrule
  \end{tabular}
  \caption{Direct comparison via Pearson's correlation (*100) between
    original AIT annotations and automated annotations from various
    annotation approaches: Rating scales, Rating scales tuples,
    Best--worst Scaling (BWS), and Paired Comparisons (PC).}
  \label{tab:results_direct}
\end{table*}

\begin{table}
  \centering\small
  \begin{tabular}{l ccccc}
    \toprule
    & Original & RS & RS-T & BWS & PC\\ 
    \cmidrule(lr){2-2}\cmidrule(lr){3-3}\cmidrule(lr){4-4}\cmidrule(lr){5-5}\cmidrule(lr){6-6}
    \multicolumn{1}{c}{Emotion}
    & AIT & D-10 & D-10  & $2N$ & 220P \\
    \cmidrule(r){1-1}\cmidrule(lr){2-2}\cmidrule(lr){3-3}\cmidrule(lr){4-4}\cmidrule(lr){5-5}\cmidrule(lr){6-6}
    Joy                  & 78.8 & 64.2 & 67.9 & 76.9 & 60.7 \\
    Anger                & 78.9 & 68.5 & 71.0 & 71.5 & 60.3 \\
    Fear                 & 78.7 & 65.4 & 64.3 & 70.9 & 53.3 \\
    Sadness              & 75.1 & 62.1 & 63.9 & 72.1 & 55.7 \\
    \cmidrule(lr){2-2}\cmidrule(lr){3-3}\cmidrule(lr){4-4}\cmidrule(lr){5-5}\cmidrule(lr){6-6}
    Mean                 & 78.3 & 65.5 & 67.1 & 73.5 & 58.1 \\ \bottomrule
  \end{tabular}
  \caption{Indirect downstream comparison via training a
    RoBERTa model on the annotated data from various
    annotation approaches: Rating scales (RS), Rating scales tuples RS-T), Best--worst Scaling (BWS),
    and Paired Comparisons (PC).}
  \label{tab:results_indirect}
\end{table}
\paragraph{Rating Scales.}
We vary the scale in the prompting method by granularity and
description. We refer to an instruction that does not contain
descriptions as ``Bare'' (B), instructions that only contain
descriptions at the maximal and minimal level as ``Outlined'' (OL),
and with complete descriptions as ``Descriptive'' (D)\footnote{The
  exact descriptions can be seen in Table \ref{tab:prompts}.}. We
experiment with intervals of 0.0--1.0 (X-1), 0--10 (X-10) and 0--4
(X-4). The latter corresponds to the rating scale used by
\cite{BWS-reliable} to compare BWS and rating scale annotations done
by humans, clipped to positive values.

\paragraph{Rating Scales Tuples.}
For each prompt we ask the model to rate four texts. Each text appears
in only one prompt. Otherwise, the approach is identical to our rating
scales approach.

\paragraph{Paired Comparisons.}
In the PC setup, we are not able to create the entirety of $N^2$
comparisons for reasons of resource constraints. We use a subset of
200 randomly selected texts per emotion to test if vastly increasing
the number of comparisons per text improves annotation quality. In
this case, direct annotation comparison only consider these 200
texts. When training a regressor, we also only use these 200 annotated
texts. In that latter case, the evaluation is performed, as in all
other regressor evaluations, on the independent test corpus.

\subsubsection{Results}
Table~\ref{tab:results_direct} shows direct evaluation results and
Table~\ref{tab:results_indirect} shows indirect downstream
results.\footnote{The annotations can be found at
  \url{https://www.uni-bamberg.de/en/nlproc/resources/autobws/}}

\paragraph{Direct Comparison.} For the direct comparison of the
annotated data, we consider
Table~\ref{tab:results_direct}. We see results for different setups of the
RS, the RS-T, the BWS results and the PC.  The results vary dramatically between
different rating scales. The best performance is achieved with the
D-10 model, which could be considered unsurprising as it provides the
most detail to the LLM. Removing part or all of the descriptions leads
to a performance drop; the values alone appear to not be
interpretable. Interestingly, changing only the scale and keeping the
descriptions (B-1/OL-1 vs.\ B-10/OL-10) also leads to a substantial
performance decrease. Apparently, floating numbers are less
informative to the model than natural numbers.

The strong difference between rating scales does not hold with
RS-T. D-10 still performs the best. B-10 shows the best results for
anger, fear and sadness, but its poor performance on joy drags its
mean score below D-10.

On all four emotions, the BWS scores are higher than any RS or RS-T
annotation. Joy and sadness show the most similar scores to the
original annotations. Similar to performance differences of human
annotators, anger appears to be most challenging.

The PC performance scores are on a similar level as BWS, but
lower. This comes, however, at a substantially higher cost: PC uses
roughly six times the annotations as BWS for only 200 texts.  Note
that this result is achieved on a different data subset.

BWS creates the annotations with the most similar performance scores
to the original annotations. However, one might argue that this is not
surprising given the alignment of the annotation method with the
original corpus creation approach. Therefore, it is important to
consider the indirect comparison.

\paragraph{Indirect Downstream Comparison.} The indirect evaluation
results, through training regressors, are shown in
Table~\ref{tab:results_indirect}. The results labeled with ``Original'' stem
from the model trained on the original human data annotations.

The RoBERTA models trained on human annotations perform well. The
results are en par with the winning team's approach in the AIT-2018
shared task \citep{duppada-etal-2018-seernet}.
BWS outperforms RS, RS-T, and PC, but not the models trained on the
original data. The drop in performance to training on the original data
is lower for joy and sadness than for anger and fear, with the latter
performing the worst. These observations are in line with
\newcite{AIT18}.

The RS and RS-T performances are similar to their direct evaluation
counterparts, though the gap in performance between emotions is
smaller. Anger is not the most challenging emotion and nearly performs the same as 
BWS. The paired
comparisons perform worse, due to the smaller training set.

\paragraph{Summary.}
BWS performs the best for both evaluation setups. It does use twice
the annotations as rating scales ($2N$ vs.\ $N$), however the increase
in performance is worth the cost, given that all annotations are
automated. Paired comparisons are too inefficient to be considered a
viable alternative.

\subsection{RQ 2: Can automated annotations be as good as human
  annotations?}
In the previous experiment, we kept the annotation setup close to the
original setup of BWS annotations to understand the impact of the LLM
use. In this section, we will exploit the advantages of automatic
annotations to see if scaling it up can improve the predictions to be
closer to human performance.

\subsubsection{Experiment Settings Details}
For the AIT annotations each tuple was annotated by three
annotators. However, running tuple sets through GPT three times is not
equivalent to annotating with three annotators; while the temperature could be raised to increase randomness in the model's output, we do not equate this to the new perspectives, experiences, intuitions, and opinions an additional annotator would provide. Hence we need another method of increasing our total number of annotations. 
The quality of BWS annotations can be increased by either increasing the number of annotators or the number of tuples \citep{BWS-reliable}. Therefore we
increase the number of tuples annotated. We start with $6N$ tuples to match the number of annotations done by AIT, which uses $2N$ unique tuples
and each tuple is annotated by 3 annotators, giving a total $6N$
annotations. Then we explore half and double that number. We only run each prompt-tuple once, but we increase the
number of unique tuples:
\begin{compactitem}
\item $3N$: 50\% more unique tuples, but half the total number of annotations.
\item $6N$: 200\% more unique tuples with the same number of total annotations.
\item $12N$: 600\% more unique tuples and twice the number of total annotations.
\end{compactitem}
The sets of tuples used for this experiment are randomly created using
the same design as AIT, but they do not purposefully contain the
original tuples.

\begin{table}
  \centering\small
  \setlength{\tabcolsep}{3pt}
  \begin{tabular}{l cccc cccc c}
    \toprule
     & \multicolumn{4}{c}{Direct Eval.} & \multicolumn{5}{c}{Ind.\ Downstream Eval.}\\
    \cmidrule(lr){2-5}\cmidrule(){6-10}
    & \multicolumn{4}{c}{GPT-3.5} & \multicolumn{4}{c}{GPT-3.5} & Orig.\\
    \cmidrule(lr){2-5}\cmidrule(){6-9}\cmidrule(rl){10-10}
    E & $2N$ & $3N$ & $6N$ & $12N$ & $2N$ & $3N$ & $6N$ & $12N$ & $2N$  \\
    \cmidrule(lr){2-5}\cmidrule(){6-9}\cmidrule(rl){10-10}
    J  & 81.0 & 78.7 & 80.5 & 81.3 & 76.9 & 76.0 & 75.7 & 77.6& 78.8 \\
    A  & 74.5 & 74.2 & 74.9 & 76.1 & 71.5 & 71.7 & 72.2 & 73.2& 78.9 \\
    F  & 76.2 & 74.3 & 76.3 & 77.3 & 70.9 & 72.6 & 73.5 & 71.7& 78.7 \\
    S  & 80.3 & 77.4 & 79.7 & 80.8 & 72.1 & 74.5 & 74.4 & 74.5& 75.1 \\
    \cmidrule(r){1-1}\cmidrule(lr){2-5}\cmidrule(){6-9}\cmidrule(rl){10-10}
    $\varnothing$  & 78.1 & 76.2 & 78.0 & 78.9 & 73.5 & 74.2 & 74.4 & 74.8& 78.3 \\
    \bottomrule
  \end{tabular}
  \caption{Evaluation with higher tuple counts.}
  \label{tab:rq2results}
\end{table}

\subsubsection{Results}
We show all results in Table~\ref{tab:rq2results}, for both a direct
comparison of the annotation output and the performance of a trained
RoBERTA-based regressor. We see that increasing the tuple counts does
lead to an improvement for both evaluations, with 12N performing best
for every emotion except fear. This follows the findings by
\citet{BWS-reliable} that increasing total annotation counts improve
annotation quality. Regarding the indirect evaluation with a trained
model, we also see an increase of performance for higher tuple
counts. With 2N, we see an average performance of 73.5 in contrast to
78.3 for humans. These gaps shrink with larger tuple counts, but not
too dramatically -- the best result is achieved with $12N$ tuples,
leading to 74.8 correlation.

\section{Further Analyses}
\subsection{Generative Model Comparison}
\label{sec:add}
We performed all experiments with GPT3.5, but the results might not
carry over to other models, and they might not remain replicable if
the API, the model, or the licenses change. We therefore compare the
results to GPT-3-davinci \citep{GPT3} and Llama2
\cite{touvron2023llama} with 13B parameters.
We use the same evaluation setup as above, with the original $2N$
tuples. Table~\ref{tab:modelsresults} shows the results.  GPT3.5
outperforms GPT3 by 14pp and Llama2 by 27pp. There is a notable drop
in performance for fear (24pp,36pp). GPT-3 performs
decently on joy and sadness (7pp/8pp drop).

\begin{table}
  \centering\small
  \setlength{\tabcolsep}{3pt}
  \begin{tabular}{l ccc ccc}
    \toprule
    &\multicolumn{3}{c}{Direct Eval.}  &\multicolumn{3}{c}{Ind.\ Downstream Eval.}\\
    \cmidrule(lr){2-4}\cmidrule(l){5-7}
    Emo. & GPT3.5 & GPT3 & Llama2 & GPT3.5 & GPT3 & Llama2 \\
    \cmidrule(r){1-1} \cmidrule(lr){2-4}\cmidrule(l){5-7}
    J & 81.0 & 73.9 & 52.2 & 76.9 & 71.7 & 68.0\\
    A & 74.5 & 61.5 & 52.1 & 71.5 & 65.9 & 49.4\\
    F & 76.2 & 51.9 & 39.5 & 70.9 & 54.7 & 49.4\\
    S & 80.3 & 71.9 & 64.1 & 72.1 & 64.7 & 62.7\\
    \cmidrule(r){1-1} \cmidrule(lr){2-4}\cmidrule(l){5-7}
    Avg. & 78.1 & 64.6& 51.5 & 73.5 & 65.3& 58.6 \\
    \bottomrule
  \end{tabular}
  \caption{Evaluation across models.}
  \label{tab:modelsresults}
\end{table}

These results translate closely to the indirect comparison. GPT-3's
regressions' performance lines up with its direct comparison
results. Llama2 does however perform better than in the direct
comparison. Joy and fear all close the gap between their Llama2 and
GPT-3 performances.

Llama2 does not perform well in our experiments. The results for joy
and sadness are acceptable, but the results for anger and fear are
even less than the paired comparisons trained on only 200 texts. The
score for joy of 68.0 is especially surprising given the low score in
the direct evaluation. This highlights that lower correlation to the
original annotations does not guarantee worse performance as a
training set. The exact cause of this is worth further investigations.
It is noteworthy that Llama2 provided further challenges, in addition
to its lower performance. Nearly 10\% of all prompts sent to the model
returned non-acceptable answers.

\begin{table}
  \centering\small
  \renewcommand{\arraystretch}{0.7}
   \begin{tabularx}{\linewidth}{lXp{2mm}p{1mm}}
     \toprule
     \multicolumn{4}{p{0.95\linewidth}}{\textbf{Ex. 1:} \textbf{Not giving a fuck is better than revenge.}\par\mbox{}\hfill\textbf{AIT: .63}  \textbf{GPT3.5: .06}} \\
     \cmidrule(r){1-2}\cmidrule(l){3-4}
     && M & L  \\
     \cmidrule(rl){3-3}\cmidrule(l){4-4}
     \multirow{12}{*}{\rotatebox{90}{Tuple 1}}
     & Yay bmth canceled Melbourne show fanfuckingtastic just lost a days pay and hotel fees not happy atm \#sad \#angry
     & 1\par 2\par 3\par G & \\
     \cmidrule(r){2-2}\cmidrule(l){3-4}
     & Just saw lil homie @NICKMERCS rage on cam. Weren't roids a thing in the late 90's or has it come back? I'm lost...
     & & 2 \\
     \cmidrule(r){2-2}\cmidrule(l){3-4}
     & Not giving a fuck is better than revenge.
     & & G \\
     \cmidrule(r){2-2}\cmidrule(l){3-4}
     & ESPN just assumed I wanted their free magazines
     & & 1\par 3 \\
     \cmidrule(){1-4}
     \multirow{14}{*}{\rotatebox{90}{Tuple 2}}
     & @MMASOCCERFAN @outmagazine No offense but the only way this makes sense is if you work for the magazine. Otherwise,who are you apologizing
     & & 1 \\
     \cmidrule(r){2-2}\cmidrule(l){3-4}
     & She's foaming at the lips the one between her hips @realobietrice, one of many great lyrics
     & 1\par 2\par G & 3 \\
     \cmidrule(r){2-2}\cmidrule(l){3-4}
     & the bee sting still suck i feel sick 
     & &  \\
     \cmidrule(r){2-2}\cmidrule(l){3-4}
     & Not giving a fuck is better than revenge.
     & 3 & 2\par G \\
     \bottomrule
   \end{tabularx}
   \caption{Example with the highest difference between manual and
     automatic annotation, and the associated tuples. The 1/2/3 refer
     to the AIT annotator IDs and G to the annotator GPT3.5.}
  \label{tab:examples}
\end{table}

\subsection{Error Analysis}
To provide an intuition why annotations differ, we manually inspect the top 10 instances per emotion that have
the largest absolute difference in emotion intensity annotation
between the human and the automatic annotation. We show these
instances in the Appendix, Table \ref{tab:topten}.

Out of the 40 texts, 21 explicitly mention the target emotion, while
19 either only refer implicitly to the emotion and could be considered
neutral, or do not describe the emotion at all. In the cases in which
the emotion is explicitly mentioned, GPT-3.5 tends to assign a higher
score than the human annotators -- in such cases, it does more often
rate the instance as the most intense (14/21 cases). In cases in which
one might argue that the text is in fact comparably neutral, GPT
assigns lower values than humans (13/19 cases). In summary, our error
analysis shows that GPT has a tendency to make consistent decisions
for explicitly mentioned emotions, but humans might have a tendency to
interpret the text, unsurprisingly, more carefully regarding implicit
information.

For reasons of space constraints in this paper, we cannot show all
tuples for all these instances. We do, however, believe that a more
in-depth error analysis requires such analysis. We resort to showing
the one example that has the highest difference in prediction for the
emotion anger (Table~\ref{tab:examples}). This instance (``Not giving
a fuck is better than revenge'') contains a strong metaphorical
negative expression. One might however argue, that it does not in fact
express anger -- it offers some freedom for interpretation. In the
first tuple all human annotators agree regarding a different instance
exhibiting most anger. GPT3.5 assigns it the lowest anger. The picture
is less clear in the second tuple. Annotators are more distributed
across instances. This analysis suggests that more combinations with
more varied tuples can lead to more reliable results. In this
instance, the error decreases from .56 to .47 when increasing the
tuple count to $12N$.

\subsection{Task Validation on Another Corpus}
\paragraph{Setup.} To validate our method's performance on emotion
intensity prediction we apply our method to a second dataset,
\textit{SemEval-2007 Task 14: Affective Text} dataset
\citep{strapparava-mihalcea-2007-semeval}, which is comprised of 1250
news headlines annotated for six emotions: joy, anger, fear, sadness,
disgust, and surprise.  The annotations were done by six annotators
using rating scales of 0--100, with each text annotated for all
emotions at once. The inter-annotator agreement score was found by
taking average Pearson's correlation scores between annotators (shown
in Table~\ref{tab:semeval_results} on the right).

We test two RS-T scales to compare with BWS $2N$: D-10, our previously
best performing scale, and OL-100, adjusted to from OL-10 to match the
dataset's label range. The experiment setup and prompts are the same
as our previous experiments. We refer to this setup as \textit{Basic
  Approach}. To further test if an annotation setup that is closer to the
original annotation environment further improves the result; we have
each prompt annotate all six emotions at once in an adapted setup
(\textit{Adapted Appr.}).

\paragraph{Results.}  Table~\ref{tab:semeval_results} shows the
results. D-10 outperforms OL-100 on every emotion and performs better
than BWS on anger, fear, and disgust in the basic approach. BWS does
better for joy, sadness, and surprise, giving BWS a better overall
performance. RS-T and BWS annotations score higher than the average
human annotator for all emotions except surprise.  In the adapted
approach, BWS scores do not change substantially, but RS-T results
improve. D-10 performs better than BWS overall and OL-100 performs en
par with BWS.

\paragraph{Interpretation.} These results
validate the method's ability to do emotion intensity prediction,
however they challenge our initial finding that BWS is the better
approach. Our interpretation is that the similarity between the
automatic annotation setup and the original setup matter -- the label
distributions substantially differ: Where rating scales allow for all
annotations to be skewed towards specific areas of the scale, BWS's
comparative nature forces scores towards a normal distribution. While
this can be a benefit when annotating fresh data, it limits BWS's
ability to replicate rating scale annotations. SemEval-2007's original
annotations are all skewed towards the low end of the scale.  Our RS-T
annotations are also skewed, but the BWS annotations are
normally distributed.

We take this as an indicator that BWS is a better choice for
annotating novel corpora from scratch automatically.
\citet{BWS-reliable} show that BWS produces more reliable annotations
than rating scales. So if our method can replicate both BWS and rating
scale annotations to a similar degree, then it follows that we should
choose the approach which performs better overall. Furthermore, when
simulating existing data, the gap in performance between BWS and RS-T
is much larger for the BWS-native dataset than for the RS-native
dataset.

\begin{table}
  \centering\small
  \setlength{\tabcolsep}{3pt}
  \begin{tabular}{l ccc  ccc  c}
    \toprule
    &\multicolumn{3}{c}{Basic Approach}  &\multicolumn{3}{c}{Adapted Approach} & Original\\
    \cmidrule(r){2-4}\cmidrule(lr){5-7}\cmidrule(l){8-8}
    &\multicolumn{2}{c}{RS-T}  & BWS &\multicolumn{2}{c}{RS-T}  & BWS & RS\\
    \cmidrule(r){2-3}\cmidrule(lr){4-4}\cmidrule(lr){5-6}\cmidrule(lr){7-7}\cmidrule(l){8-8}
    Emo. & B-100 & D-10 & $2N$ & B-100 & D-10 & $2N$ & B-100\\
    \cmidrule(r){1-1} \cmidrule(lr){2-3}\cmidrule(lr){4-4}\cmidrule(lr){5-6}\cmidrule(lr){7-7}\cmidrule(l){8-8}
    Joy  & 47.7    & 66.8 & 70.4   & 71.2 & 76.2 & 70.1 & 59.9\\
    Ang. & 44.3    & 60.1 & 59.6   & 57.1 & 61.0 & 60.2 & 49.6\\
    Fear & 65.1    & 68.4 & 65.8   & 66.2 & 72.9 & 67.5 & 63.8\\
    Sad. & 66.5    & 71.5 & 74.2   & 73.4 & 77.3 & 73.0 & 68.2\\
    Dis. & 33.1    & 47.8 & 47.5   & 51.4 & 52.7 & 49.4 & 44.5\\
    Sur. & 18.5    & 15.6 & 32.3   & 21.5 & 23.8 & 24.6 & 36.1\\ 
    \cmidrule(r){1-1} \cmidrule(lr){2-3}\cmidrule(lr){4-4}\cmidrule(lr){5-6}\cmidrule(lr){7-7}\cmidrule(l){8-8}
    Avg. & 57.2    & 67.5 & 68.4   & 68.0 & 72.9 & 68.5 & 53.7\\
    \bottomrule
  \end{tabular}
  \caption{Results for applying our method to SemEval-2007 Task 14 dataset. Basic
  approach uses the same prompts as previous experiments. Adapted approach 
  rates all emotions in a single prompt. Original shows the average 
  inter-annotator agreement score of the 6 human annotators.}
  \label{tab:semeval_results}
\end{table}

\section{Conclusion \& Future Work}
We proposed to automate annotations of text data with continuous
labels with BWS, which outperforms rating scales and paired
comparisons in the case of emotion intensity predictions, when the
original annotations were also annotated using BWS. The predictions
from a regression model, fine-tuned on automated annotations, perform
nearly en par with the models fine-tuned on the original human
annotations. We showed that we can improve the annotation quality by
increasing the total number of tuples.  In general, we conclude that
BWS is the better approach to annotate novel data sets for emotion
intensity regression.

The results of our experiments are encouraging for emotion intensity
regression. We presume that these findings carry over to other
regression tasks, but this still needs to be validated. Candidates for
other tasks would be the BWS-labeled toxicity data set Ruddit
\citep{hada-etal-2021-ruddit} or the Affective Norms for English Words
dataset \citep{bradley1999affective}. Word similarity assessment tasks
could be an interesting case for evaluation as well
\citep{antoniak-mimno-2018-evaluating}.

Finally, it is important to study more open-source generative models,
instead of relying on pay-locked and black-box models.

\section*{Acknowledgements}
We thank Saif Mohammad for helping us and providing us the original
tuples of the data set we used. This project is partially supported by
the project ITEM (User’s Choice of Images and Text to Express Emotions
in Twitter and Reddit, funded by the German Research Foundation, KL
2869/11-1). We thank the reviewers and the action editor at ACL
Rolling Review for their helpful feedback.

\section{Limitations}
Before our method can be a reliable alternative to manual annotations
it must be tested on more NLP tasks. While the results are promising
on predicting emotion intensity, several possible shortcomings come to mind. Without
manually annotated data to compare to, it is difficult to judge the
quality of the automated annotations. Especially for novel tasks, it
is hard to judge if mediocre performance of the regression model is
evidence of poor quality annotations or of the task being difficult to
model.

The results we present on the task of predicting emotion intensity
rely on the assumption that none of the models used were pretrained on
the AIT dataset. The poor results from prompts using rating scales
leads us to believe the model does not have prior knowledge of the
dataset. This highlights the problem with using GPT-3 as it is a
black-box system which inhibits interpretability.

At the time of our experimentation, the token limit of generative
large language models posed a challenge due to the need for four texts to
be apart of each prompt. With the recent (and fast) development of
improved models, this constitutes a limitation of our research. Future
research may look into annotating longer texts.

Our method relies on the information embedded in an LLM to accurately
compare target texts. If the task is too specialized than the LLM
might require fine-tuning to perform adequately. This defeats the
purpose of the method as it is intended to circumvent the need for
training data. Furthermore, the method is not as accessible as working
with non-generative LLMs such as RoBERTa. The success found in this
paper relied on GPT models which are not open source, making them less
transparent and more unreliable.

\section{Ethical Considerations}
Our method does not contribute a new data set or introduce a
novel task. Therefore, it does not add any additional risks from these
perspectives to the already existing research landscape. All data that
we use, we use for their originally intended case, namely the creation
of emotion intensity prediction models.

However, it is noteworthy that previous research showed that emotion
analysis systems are biased for various reasons
\citep{kiritchenko-mohammad-2018-examining}. The use of language
models for automatic creation might lead to different biases and this
requires further research.

The core idea behind our method is to reduce the need for manual
annotations. Ideally, researchers would only need manual annotations
for their development and test sets, reducing manual annotations by
roughly 80\%. While this is great for research projects, this can
reduce the amount of work available for people who depend on
annotation-based jobs.

Though, the reduction could be attractive for tasks which require
annotators to read emotionally or mentally damaging texts, such as
hate speech or toxicity. Abusive content detection systems are needed
more in online spaces but the creation of such systems requires
annotators to spend time interacting with content which can cause
undue emotional trauma \citep{contentModchallenges, contentethics}.

\balance
\bibliography{anthology,custom}

\appendix
\onecolumn

\section*{Appendix: Text Examples with the Largest
  Manual and Automatic Annotation Difference}

\begingroup
  \centering\small
  \renewcommand{\arraystretch}{0.8}
\begin{tabularx}{\linewidth}{lXccc}
\toprule
  Emo. & Text & AIT & GPT    & $\Delta$ \\
  \cmidrule(r){1-1}\cmidrule(lr){2-2}\cmidrule(lr){3-3}\cmidrule(lr){4-4}\cmidrule(l){5-5}
\multirow{13}{*}{\rotatebox{90}{Joy}} & \#LethalWeapon A suicidal Vet with PTSD... so FUCKING FUNNY....   let the hilarity begin...    & .64  & .06 & -.58 \\
   & It was very hard to stifle my laughter after I overheard this   comment. It really is amazing in the worst ways. & .65  & .12  & -.52  \\
   & @ardit\_haliti I'm so gutted. I loved her cheery disposition. & .50  & .0 & -.50    \\
   & Rojo is so bad it's hilarious.  & .60   & .13  & -.48  \\
   & I fear that if United bottle this my heart would actually   collapse from laughter.  & .69  & .25   & -.44  \\\cmidrule(){2-5}
   & I wish there were unlimited glee episodes:( so I could watch   them forever. \#gleegoodbye   & .31  & .75   & .44   \\
   & @OrbsOfJoy plan a date... like a date u find pleasing or smth.   fuckign\textbackslash{}n\textbackslash{}n10/10. because the child will grow to be a ten out of ten   & .31  & .75   & .44   \\
   & Lea doing a mini set tour of glee my heart just cried tears of   happiness and sadness    & .49  & .94 & .45  \\
   & \#RIP30 Heaven is rejoicing because they've gained an angel, the
     Keifer family are in my prayers & .40   & .88  & .48   \\
    & Headed to Montalvo w/@jaxster3—bring on the \#mirth,   bitches!\textbackslash{}nd(-\_-)b\textbackslash{}n@Nick\_Offerman\textbackslash{}n@MeganOMullally\textbackslash{}n\#SummerOf69Tour2016    & .44    & .94 & .50  \\\cmidrule(){1-5}
\multirow{18}{*}{\rotatebox{90}{Anger}}   & Not giving a fuck is better than revenge.  & .63   & .06 & -.56 \\
    & @FluDino Event started! everyone is getting ready to travel to   the lake of rage, where everything glows    & .52   & .0  & -.52  \\
    & could never be a angry drunk lol yall weirdos just enjoy your   time  & .52   & .0 & -.52  \\
    & @Iucifaer you can go on what you usually do its just their own   personal reason and not mean to offend anyone :(   & .63   & .13  & -.50    \\
    & Inner conflict happens when we are at odds with ourselves. Honor   your values and priorities.    \#innerconflict \#conflict \#values  & .50 & .0  & -.50    \\\cmidrule(){2-5}
    & The war is right outside your door \#rage \#USAToday   & .50 & .94 & .44  \\
    & Marcus Rojo is the worst player i have ever seen. Useless   toasting burning bastard  & .56   & 1.0  & .44   \\
    & @Bell @Bell\_Support Cancelling home Fibe, Internet and TV this   afternoon - as soon as I can arrange alternate Internet. 2/2 \#angry \#fedup     & .48   & .94 & .45  \\
    & You boys dint know the game am I the game... life after death...   better chose and know who side you on before my wrath does come upon us & .52   & 1.0  & .48   \\
    & And Republicans, you, namely Graham, Flake, Sasse and others are   not safe from my wrath, hence that Hillary Hiney-Kissing ad I saw about you     & .35   & .88  & .52   \\\cmidrule(){1-5}
\multirow{18}{*}{\rotatebox{90}{Fear}}    & @RJAH\_NHS @ChrisHudson76 @mbrandreth \#course day \# potential   Leadership \#excited \#nervous \# proud    & .65   & .13  & -.52  \\
    & I was literally shaking getting the EKG done lol & .88   & .38  & -.50    \\
    & MSM stoking \#fear. Please remember the beautiful prayerful   protests in Dallas and Atlanta. Smile at a stranger. We make each other   strong.    & .56   & .06 & -.50 \\
    & @ChrissyCostanza and have social anxiety. There is many awkward   things wrong with me.  & .77   & .31 & -.46 \\
    & Avoiding \#fears only makes them scarier. Whatever your \#fear, if   you face it, it should start to fade. \#courage    & .71   & .25   & -.46  \\\cmidrule(){2-5}
    & Staff on @ryainair FR1005. Asked for info and told to look   online. You get what you pay for. \#Ryanair @STN\_Airport \#Compensation \#awful  & .27   & .75   & .48   \\
    & I'm mad at the injustice, so I'm going to smash my neighbours   windows'.  Makes perfect sense.  \#CharlotteProtest \#terrible  & .46   & .94 & .48  \\
    & O you who have believed, fear Allah and believe in His   Messenger; He will {[}then{]} give you a double portion of His mercy...' (Quran   57:28)  & .33   & .88  & .54   \\
    & Don't think I'll hesitate to run you over. Last time I checked,   I still had 'Accident Forgiveness' on my insurance policy...  & .39   & .94 & .55  \\
    & @dc\_mma @ChampionsFight think shes afraid to fight Holly. One   can only imagine what goes through her head when she thinks of Cyborg \#terror    & .40   & 1.0  & .60   \\\cmidrule(){1-5}
\multirow{14}{*}{\rotatebox{90}{Sadness}} & It feel like we lost a family member & .71   & .19 & -.52 \\
    & @chelseafc let them know it's the \#blues & .52   & .06 & -.46 \\
    & It's a gloomy ass day & .89 & .44 & -.45 \\
    & @Theresa\_Talbot @fleurrbie Haha...sorry about the dreadful   puns... I need to get out more....I've been cooped up lately... & .56   & .13  & -.44  \\
    & @Beakmoo hmmmm...you may have a point... I thought Twitter had   got dull. LAMINATION     & .54   & .13  & -.42  \\\cmidrule(){2-5}
    & So unbelievably discouraged with music as of late. Incredibly   behind on Completing my album. Not digging this at all.     & .60   & 1.0  & .40   \\
    & Nothing else could possibly put a damper on my day other than   doing X-rays on someone with kickinnnnn ass breath  & .40   & .81 & .41  \\
    & @LBardugo Start w/ the 3 songs in Blue Neighborhood\textbackslash{}n1)   Wild\textbackslash{}n2)Fools\textbackslash{}n3)Talk Me down for \#Wesper\textbackslash{}nAlso,\textbackslash{}n4)Too   Good.  \#serious kaz/inej feelz & .38   & .81 & .43  \\
    & @ticcikasie1 With a frown, she let's out a distraught   'Gardevoir' saying that she wishes she had a trainer & .48 & .94 & .46  \\
 & @courtneymee I'm 3 days sober don't wanna ruin it & .33   & .81 & .48 \\ \cmidrule(){1-5}
\end{tabularx}
\captionof{table}{\label{tab:topten} Top ten texts with largest errors per emotion for GPT-3.5 annotations compared to original AIT annotations.}
\endgroup

\end{document}